\documentclass[letterpaper, 10 pt, conference]{ieeeconf}  % Comment this line out if you need a4paper

\IEEEoverridecommandlockouts                              % This command is only needed if 
\usepackage{amsmath} % assumes amsmath package installed
\usepackage{amssymb}  % assumes amsmath package installed
\usepackage{eucal}
\usepackage{booktabs}
\usepackage{graphicx}
\usepackage{subcaption}
\usepackage{hyperref}
\usepackage{algorithm}
\usepackage{algpseudocode}
\usepackage{xcolor}
\usepackage{hyperref}
\usepackage{caption}
\usepackage{bm}
\usepackage{multirow}
\usepackage{pifont}% http://ctan.org/pkg/pifont
\newcommand{\cmark}{\ding{51}}%
\newcommand{\xmark}{\ding{55}}%
\newcommand*\Update{\color{black}}
\newcommand*\Done{\color{black}}

\title{\LARGE \bf
PhyPush: One Push is All You Need \\ for Sensorless Physical Property Estimation \\ with Physics-Guided Transformers
}

\author{
    Koyo Fujii$^{1}$, Luis Figueredo$^{1}$, Praminda Caleb-Solly$^{1}$, Ivan Boschi$^{2}$, \\
    Edoardo Ida'$^{2}$, Marco Carricato$^{2}$, and Aly Magassouba$^{1}$
    \thanks{$^{1}$ School of Computer Science, University of Nottingham, Nottingham, United Kingdom. $^{2}$ Dept. of Industrial Engineering, University of Bologna, Bologna, Italy.
    \Update
    Author's Accepted Manuscript. Released under the Creative Commons license: Attribution 4.0 International (CC BY 4.0). Corresponding author: Koyo Fujii, email: koyo.fujii@nottingham.ac.uk
    \Done
    }
}

\begin{document}

\maketitle
\thispagestyle{empty}
\pagestyle{empty}

\begin{abstract}
Accurately estimating object mass and friction is fundamental to achieving reliable and adaptive robotic manipulation. Although interactive perception provides a powerful mechanism for inferring such properties, most existing approaches depend on specialized hardware such as force/torque sensors, tactile arrays, or multi‑camera motion‑capture systems, limiting scalability and deployment. This paper introduces PhyPush, a physics‑guided Transformer that estimates an object’s mass and friction coefficient using the end‑effector velocity from a single push. PhyPush incorporates constraints from Newton’s second law and the Coulomb friction model through a physics-guided loss, improving physical consistency and generalization to unseen objects and surfaces.
Across diverse simulation and real‑world setups, PhyPush consistently achieves more accurate  estimation. In simulation, it reduces error by over 10\% compared with a baseline that has privileged access to full force information in out of domain conditions, which are further validated in zero-shot real-world experiments.
Overall, the results demonstrate that physics‑guided learning can enable low‑cost, sensor‑efficient estimation of physical properties, relying solely on a single push and readily available kinematic data.
Our implementation is publicly available at \url{https://github.com/koyo-8144/PhyPush_IROS2026}.

% Extensive simulation and real‑world experiments in various configurations demonstrate that PhyPush achieves  more accurate mass and friction estimation in challenging out‑of‑domain scenarios, reducing error by more than 10\% compared to 
% a baseline with privileged information and direct access to full force information. 
% baseline methods. 
% These results highlight the potential of physics‑guided learning to enable low‑cost, sensor‑efficient physical property estimation for robotic manipulation with readily deployable setups based on a single push and standard kinematic measurements alone.
\end{abstract}

\section{INTRODUCTION}

Understanding the physical properties of objects is essential for embodied AI systems that must interact in the real world safely and reliably.
In real-world settings, robots are expected to manipulate various objects, often with unknown physical properties, under unstructured conditions.
This uncertainty makes complex manipulation difficult: the outcomes of contact-rich actions are hard to predict, increasing the likelihood of task failure or even harm to nearby humans.
Key interaction outcomes depend critically on the object's properties, such as mass and frictional characteristics.
Although visual cues can provide coarse human-like intuition about these properties \Update \cite{standley_image2mass_2017, gao_physically_2024} \Done
these latent properties cannot be fully inferred from vision alone.
% gao2024physically zhai_physical_2024 andrade_improving_2023

\begin{figure}[t]
    \centering
    \includegraphics[width=\columnwidth]{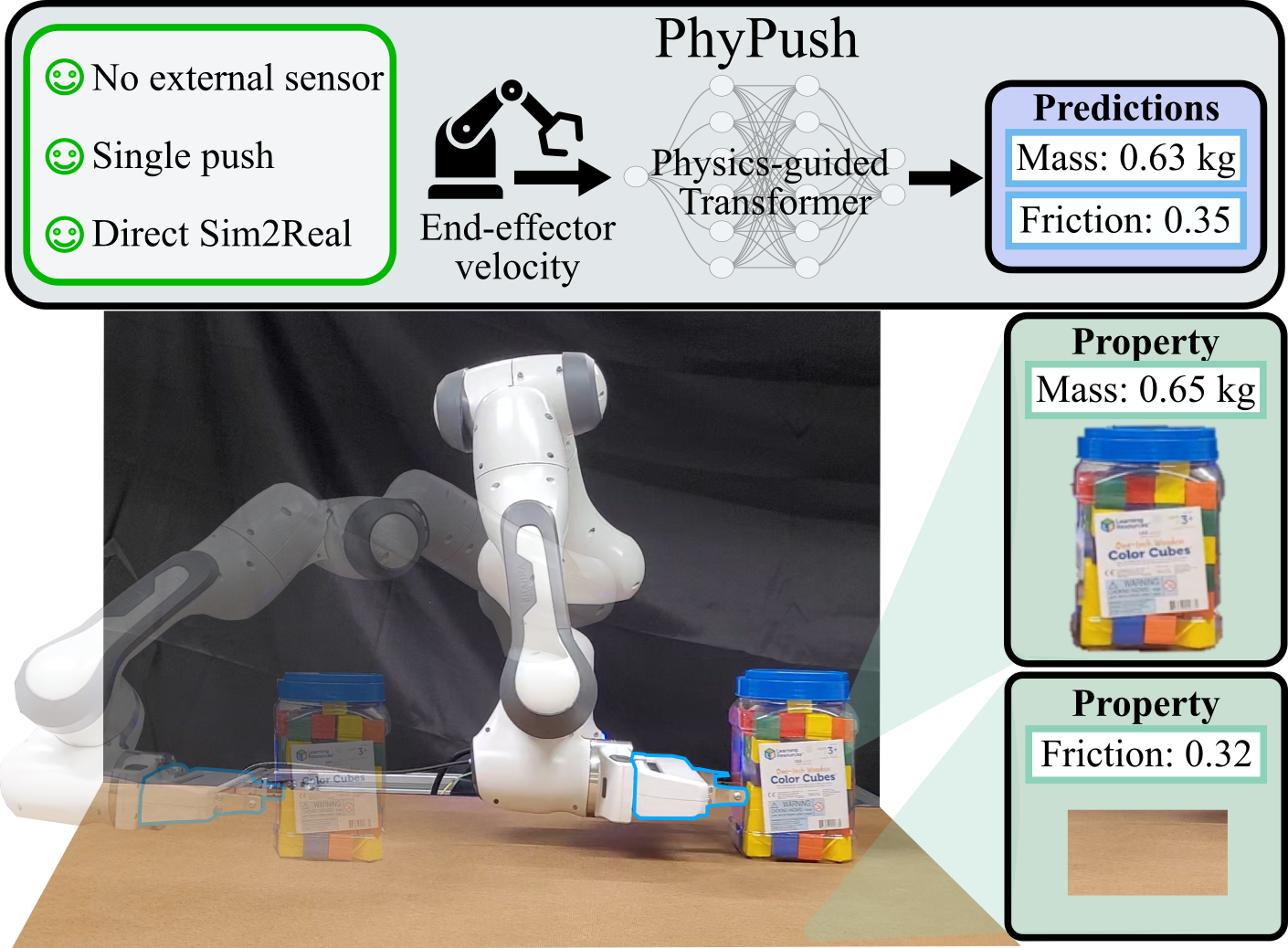}
    \caption{
        \small 
        One push is all PhyPush needs: from a single translational interaction, our physics-guided transformer framework predicts an object’s mass and friction using only the end-effector velocity profile, avoiding specialized sensing hardware such as force/torque sensors, tactile sensing, and external motion-capture systems.
        % Our PhyPush estimates an object's mass and friction coefficient precisely based on only a kinematic end-effector velocity from a single push without expensive F/T or tactile sensors or complex camera setups. 
    }
    \label{fig:eye_catch}
\end{figure}

To bridge this gap, prior work has explored various low-risk active interactions, such as pushing~\cite{mavrakis_estimating_2020} and pulling~\cite{dutta_predictive_2025}, lifting, grasping, or poking~\cite{kruzliak_interactive_2024} to uncover the physical parameters that enable more reliable manipulation.
However, these methods often rely on expensive and specialized hardware, such as force/torque (F/T) sensors, tactile arrays, and complex multi-camera motion capture systems. 
These setups are costly, difficult to maintain, and often impractical outside controlled laboratory environments. 
Moreover, existing techniques typically require multiple and diverse interactions to infer an object's physical properties, substantially increasing identification time and reducing overall efficiency.
Such requirements are often unsuitable for widespread deployment and make it challenging to reproduce these results in dynamic real-world environments.

To overcome these limitations, we introduce PhyPush, a physics-guided Transformer framework \cite{vaswani2017attention} designed to estimate an object's mass and friction coefficient. 
Unlike previous methods, our approach relies solely on kinematically calculated robotic end-effector velocity, available in any robotic arm, eliminating the need for specialized hardware.
Moreover, PhyPush requires only a single translational push to accurately estimate these physical properties.
The key insight behind PhyPush is that objects with different physical properties exhibit distinct dynamic responses when subjected to the same push command.
As long as the end‑effector maintains contact, its motion becomes tightly coupled to the object’s behavior, making the velocity profile a rich source of information for physical property estimation.

PhyPush leverages a Transformer-based temporal model trained with physics-guided loss.
This architecture captures long-range temporal dependencies within the end-effector velocity profile, allowing the network to selectively attend to the most informative time frames for mass and friction estimation. 
Our physics‑guided loss, derived from Newton’s second
law and Coulomb’s law of friction~\cite{gagnon_review_2019}, further shapes the attention mechanism.
By grounding the learning process in these fundamental laws, the model does not merely memorize statistical correlations, unlike traditional data-driven loss.
Instead, it learns to focus on the time intervals that are physically relevant to the estimation task.
Specifically, Newtonian effects dominate the transient phase, when the object accelerates, while Coulomb friction governs the steady-state phase, when the object move at constant velocity. Building on this insight, our main contributions are:
\begin{itemize}
    \item We introduce PhyPush, a novel framework for estimating physical properties using only kinematic end‑effector velocity, removing the need for expensive F/T or tactile sensors.
    \item We develop a Transformer architecture with a physics‑guided loss that embeds constraints from Newton’s second law and the Coulomb friction model directly into training, enhancing physics fidelity and improving generalization.

    % \item We conduct extensive evaluations demonstrating accurate estimation of mass and friction coefficients, including in unseen settings where we achieve over 10\% improvement compared with a baseline method relying on an F/T sensor. 
    % Real-world experiments further validate the effectiveness of our physics‑guided loss relative to a conventional data‑driven loss.

    \item \Update We conduct extensive evaluations demonstrating accurate estimation of mass and friction coefficients, including in unseen simulation settings where we achieve over 10\% improvement compared with a baseline method relying on an F/T sensor.
    Crucially, our real-world experiments demonstrate successful zero-shot sim-to-real transfer without any fine-tuning or retraining, further validating the effectiveness of our approach. \Done
    
\end{itemize}

% We evaluated our framework
% in simulation and real-world experiments and achieved more
% accurate friction coefficient estimation for the most challenging
% out-of-domain scenarios by reducing more than 20% error
% compared to a force input required baseline.

\section{RELATED WORKS}

% Perception: 
% ref1: Estimating Material Properties of Interacting Objects Using Sum-GP-UCB

% ref2: Phys2Real: Fusing VLM Priors with Interactive Online Adaptation for Uncertainty-Aware Sim-to-Real Manipulation

% ref3: Physically Grounded Vision-Language Models for Robotic Manipulation

% ref4: Estimating Object Physical Properties from RGB-D Vision and Depth Robot Sensors Using Deep Learning

% Interactive exploration:
% ref5: Unsupervised Discovery of Objects Physical Properties Through Maximum Entropy Reinforcement Learning

% Physics approach:
% ref6: Learning Object Properties Using Robot Proprioception via Differentiable Robot-Object Interaction

% Recent work incorporates physical laws directly into the learning process. 
% \cite{wu_physics_2016} estimated mass, density, volume, and friction/restitution coefficients by observing unlabeled object interactions (e.g., sliding, falling, floating), combining deep learning with physics-informed generative models guided by physical laws. 

Physical property inference remains an active research topic in the robotics community. Several lines of research have been explored using vision-only, active interaction, or, more recently, physics-informed learning.

\textbf{Vision‑only approaches:}  These methods infer latent properties such as mass, stiffness, or friction from RGB or RGB‑D data \cite{seker2024estimating}. For example \cite{wu_galileo_nodate} estimated  mass, friction and shape by fitting simulations to observed video data using probabilistic inference, while \cite{cardoso2025estimating} predicted object mass directly from RGB-D data. More recent approaches relied on multimodal large language models \cite{gao2024physically,wang2025phys2real} leveraging commonsense priors to improve estimations. Yet because many physical properties are not directly observable, vision‑based methods often struggle with invisible parameters such as mass or friction, especially in unstructured environments. 

\textbf{Interaction-based approaches:} A second major line of work uses active interaction to obtain more informative signals about object dynamics. Pushing, in particular, is a widely studied probing action \cite{bauza2019omnipush}, as the resulting motion reveals object‑specific physical properties. These methods typically exploit force, torque, tactile, or object‑trajectory data to infer parameters such as mass, friction, or Center of Mass (CoM) \cite{chareyre2025unsupervised}. DensePhysNet \cite{xu_densephysnet_2019} followed this paradigm by learning physical properties directly from observed object motion during sliding and collisions. More recent multimodal approaches combined haptic and vision sensors \cite{mavrakis_estimating_2020, dutta_push_2023}, achieving higher accuracy than vision‑only methods. 
Other works \cite{kruzliak_interactive_2024, dutta_predictive_2025} explored richer multi‑action strategies, such as poking, shaking, or tapping, to refine estimates. Despite their effectiveness, these systems often rely on specialized hardware, controlled environments, or repeated interactions, which limits scalability and makes deployment on standard robotic arms more challenging.

\textbf{Physics-informed approaches:} A complementary line of work incorporates physical structure directly into the learning process. \cite{chen2025learning} introduced a differentiable‑physics approach that estimates the properties of objects using only proprioceptive signals of the robot joint-space during a grasping task. Although promising, the approach was evaluated on a limited set of light objects, which restricts its generalization. Moreover, differentiable simulators are computationally demanding and require long optimization cycles, making them difficult to deploy in fast and contact-rich real-world tasks.

Similarly, PhyPush belongs to the broader family of physics‑informed learning approaches \cite{cai_physics-informed_2021} that have recently gained traction in robotics~\cite{sorrentino2024physics}. However, in contrast to \cite{chen2025learning}, PhyPush relies only on kinematically computed end‑effector velocity, readily available on any robotic arm, to estimate physical properties. Physics constraints are incorporated  through loss‑function design \cite{daw_physics-guided_2022} avoiding the overhead of differentiable simulation. This lightweight physics‑guided formulation preserves the benefits of interaction‑based and physics‑guided learning while avoiding the practical constraints of specialized sensors or expensive optimization.

\section{METHODOLOGY}

\begin{figure*}[t]
    \centering
    \includegraphics[width=1\textwidth]{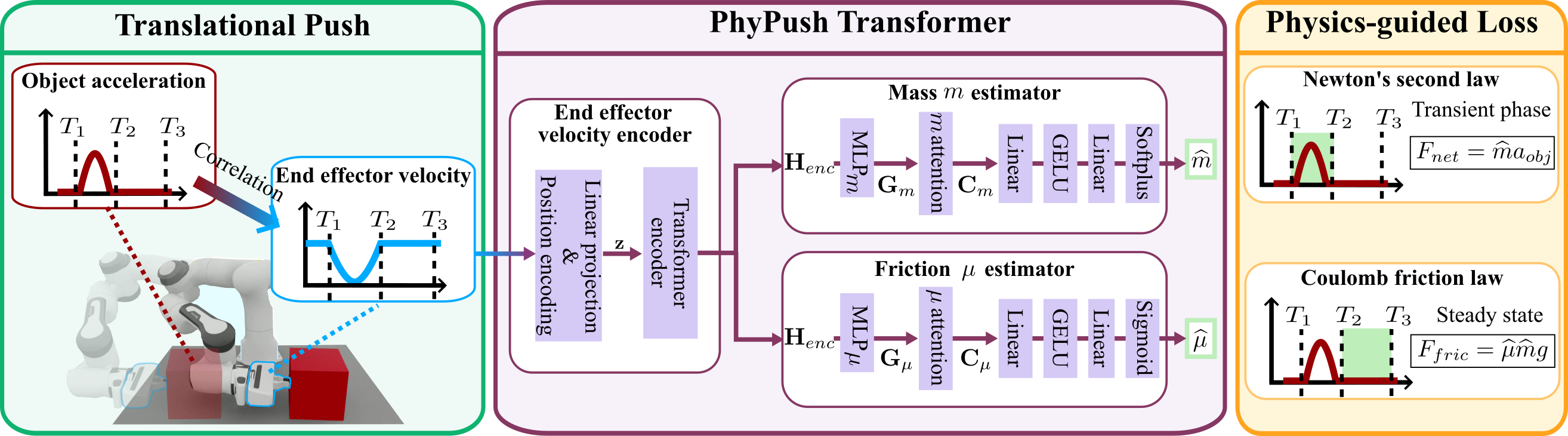}
    \caption{PhyPush system pipeline:
        \small The core intuition behind our PhyPush is that an object with distinct properties exhibits unique behavioral responses to the same push action. Therefore, our framework estimates these properties by observing the resulting object dynamics.
        Specifically, the robot applies a single translational push where the end-effector maintains continuous contact with the object, allowing the end-effector velocity to serve as a strong input signal to our estimator.
        Our estimator utilizes a Transformer architecture, consisting of the end-effector velocity encoder to process the velocity sequence, followed by independent mass and friction estimators.
        These estimations are co-optimized using physics-guided loss that integrates Newton's Second Law for the transient contact dynamics phase and Coulomb's law of friction \Update only in simulation and is directly transferred to real hardware without finetuning or retraining. \Done
        % This grounding in fundamental physical laws yields superior robustness and generalizability compared to traditional data-driven loss models.
    }
    \label{fig:pipeline}
\end{figure*}

We propose PhyPush, a physical property estimator framework, where a robot applies a single translational push to an object and estimates the object's physical properties.
PhyPush is a Tranformer-based estimator trained with physics-guided loss. This encourages physically grounded estimation by directing the model’s attention to the key parts of the end‑effector kinematic sequence.
% Our goal is to learn an estimator that determines the mass $m$ and the friction coefficient $\mu$ of an object through a single-push maneuver. 
% The robot executes a constant velocity command $v^{\text{ee, cmd}}$, and the model observes the resulting sequence of measured end-effector velocities, denoted as $\mathbf{v}_{0:T_2} = (v^{\text{ee, meas}}_0, \dots, v^{\text{ee, meas}}_{T_1}, \dots, v^{\text{ee, meas}}_{T_2})$.

\subsection{Task Definition}
We aim to estimate the mass $m$ and friction coefficient $\mu$ of an object from the end‑effector velocity generated during the push, without external sensors or cameras. 
The kinetic friction coefficient $\mu$ is defined as the ratio of the friction force $F_{\text{fric}}$ between two surfaces to the normal force $F_{\text{norm}}$ pressing them together while the object is in motion. This parameter characterizes the resistance encountered as one surface slides over another.
In our setup, a robotic arm pushes an object on a tabletop planar surface. 
We consider tabletop objects with uniformly distributed mass so that their CoM coincides with their geometric center.
% The objects are characterized by a mass range of $m \in [0.3, 1.0]$~kg and a friction coefficient range of $\mu \in [0.3, 0.5]$.
% \Update
% The objects are characterized by a mass range of $m$ $\in$ $[0.2, 3.0]$~kg, and object-table surface pairs are characterized by a friction coefficient range of $\mu$ $\in$ $[0.15, 0.7]$.
% \Done
% \textcolor{blue}{TODO: To compare with the range of previous works (if that is anyhow beneficial to us. Also, in our experiments, we don't have a purely uniform (that is the idealistic scenario, but we generalize, which is part of the problems objectives/definition. Perhaps, a formal problem defintion since it is already well-defined and impressive (in terms of simple input, and estimation output and generalization goals.}

\subsection{Pushing Strategy}
We use a single translational-pushing strategy. 
A single-push approach significantly reduces the time needed for system identification, thereby increasing the robot's overall operational efficiency.
The robot pushes an object once with its gripper in a fixed open configuration, moving directly from a pre-push position toward the object. 
% Contact occurs at two points, $p^{\text{contact},1}$ and $p^{\text{contact},2}$, between the surfaces of the open gripper and the object surface. 
To enable the object to move without lateral deviations, the push is aligned with the CoM of the object.
% such that the push point $p^{\text{push}}$ is the midpoint of the contact interface.
% This ensures consistency across different object geometries while preventing rolling or unintentional pivoting: 
% \textcolor{blue}{The $p^{\text{push}}$ can be a bit confusing as there is no actual physical point or push in the real robot. Also the centered between the points is not relevant. The relevant is that is positioned at the COM. Not sure if we also need the equation. It is rather simple. }
% \begin{equation}
%     p^{\text{push}} = \frac{p^{\text{contact},1} + p^{\text{contact},2}}{2}
% \end{equation}

The end-effector is commanded at a constant translational velocity $v^{\text{ee, cmd}}$ and maintains continuous contact with the object throughout the duration of the push.
During execution, the end-effector velocity sequence is recorded at 100~Hz. 
To synchronize the physical phases, we identify the specific time step $t = T_{\text{min}}$ where the end-effector’s acceleration reaches its highest negative value, representing the moment of contact impact.  
From this trigger point, we extract a window size of $T$
% \textcolor{blue}{This is a very adhoc number. We either explain it, or leave it open for now (as a decision variable and show only in the experiments)}
time steps of the measured end-effector's velocity $v^{\text{ee, meas}}$ at 100~Hz, spanning from $t = T_1 = T_{\text{min}}$ through the transient phase till $t = T_2 = T_{\text{min}} + \frac{T}{2}$ to the steady-state phase at $t = T_3 = T_{\text{min}} + T$. 
\Done
This processed velocity sequence serves as the primary input to our model, ensuring that the input window consistently captures both the initial inertial dynamics and the subsequent frictional sliding.

% To capture both transient and steady-state dynamics, we implement a velocity profile where the end-effector and object accelerate from $t=0$ to $t=T_1$ and maintain a constant velocity from $t=T_1$ to $t=T_2$.

\subsection{Transformer-based learning architecture}
PhyPush is an encoder-only, dual-stream Transformer designed to separately address the transient dynamics of inertia and the steady-state dynamics of friction. 
We consider a sequence of one-dimensional translational end-effector velocities recorded during the contact interval, $\mathbf{x} \in \mathbb{R}^{T \times 1}$, where $T$ denotes the sequence length, as input to our system to estimate the object's mass $\widehat{m}$ and friction coefficient $\widehat{\mu}$. 
 
% In our implementation, we utilize a velocity sequence, resulting in an input dimensionality of $d_{\text{in}}=1$.
% In our implementation, we utilize a velocity sequence of $T=60$ time frames recorded at 100~Hz, resulting in an input dimensionality of $d_{\text{in}}=1$.

\subsubsection{Input Projection and Embedding}
During the initial processing phase, $\mathbf{x} \in \mathbb{R}^{T \times 1}$ is linearly projected into a high-dimensional latent space, of dimension $d_{\text{model}} = 64$.
% , and subsequently normalized using layer normalization. 
To prevent these physical embeddings from being overshadowed by the positional values, they are scaled by the square root of the latent dimension. Finally, a fixed sinusoidal Positional Encoding ($\text{PE}$) is injected to preserve the temporal order of the sequence. This embedding process is formulated as % follows:
% \begin{equation}
%     \tilde{\mathbf{z}} = \text{LayerNorm}(\mathbf{x} \mathbf{W}_{\text{in}} + \mathbf{b}_{\text{in}})
% \end{equation}
\Done
\begin{align}
    \tilde{\mathbf{z}} = \ \mathbf{x} \mathbf{W}_{\text{in}} + \mathbf{b}_{\text{in}} \text{ and } 
    \mathbf{z} = \ \tilde{\mathbf{z}} \sqrt{d_{\text{model}}} + \text{PE}, 
\end{align}
where $\mathbf{W}_{\text{in}}$ and $\mathbf{b}_{\text{in}}$ are the projection weights and biases, and the final embedded sequence is $\mathbf{z} \in \mathbb{R}^{T \times d_{\text{model}}}$.

\subsubsection{Transformer Encoder}
The sequence $\mathbf{z}$ is processed through a standard Transformer Encoder ($\mathcal{E}$) containing multi-head self-attention and feed-forward layers. 
This allows the network to build a global contextual understanding of the entire kinematic input sequence
% \textcolor{blue}{Should we give a symbol, instead of calling the function TransformerEncoder? }
\begin{equation}
    \mathbf{H}_{enc} = \mathcal{E}(\mathbf{z}),
\end{equation}
where $\mathbf{H}_{enc} \in \mathbb{R}^{T \times d_{model}}$.

\Done
\subsubsection{Dual-Stream Feature Separation}
To isolate the underlying physical effects, $\mathbf{H}_{enc}$ is fed into two independent streams using two Multi-Layer Perceptrons (MLPs). Each branch processes the same input to produce its own latent representation ($\mathbf{G}_{m}$ and $\mathbf{G}_{\mu}$) as  
\begin{align}
    \mathbf{G}_{m} = \text{MLP}_{m}(\mathbf{H}_{enc}) \text{ and }
    \mathbf{G}_{\mu} = \text{MLP}_{\mu}(\mathbf{H}_{enc}).
\end{align}
\Done \subsubsection{Physical Property Attention}
Instead of averaging over the $T$ time steps of $\mathbf{G}_{m}$ and $\mathbf{G}_{\mu}$, our architecture applies a dual-stream attention-based pooling mechanism. Each stream uses a query $\mathbf{q}$, to selectively attend to the most informative intervals of the sequence for estimating $m$ and $\mu$, respectively. For a property $p \in \{m, \mu\}$, we use a trainable global query $\mathbf{q}_p \in \mathbb{R}^{1 \times 1 \times d_{\text{model}}}$ together with its corresponding feature $\mathbf{G}_p$. 
The query is scaled by a sharpness hyperparameter $s_p$ to produce attention weights $\mathbf{\alpha}_p$. Attention $\mathbf{\alpha}_p$,  which highlights the time intervals most relevant to the underlying physical dynamics, can be expressed as:
\begin{equation}
    \mathbf{\alpha}_p = \text{Softmax}\left(\frac{(s_p \mathbf{q}_p \mathbf{W}_Q^p)(\mathbf{G}_p \mathbf{W}_K^p)^T}{\sqrt{d_k}}\right),
    \label{eq:att_weight}
\end{equation}
where $\mathbf{W}_Q^p$ and $\mathbf{W}_K^p$ are the query and key projection matrices and $\sqrt{d_k}$ is a scaling factor. For each property $p$, the value projection matrix $\mathbf{W}_V^p$  maps each feature $\mathbf{G}$ into the value space.  The attention weights $\alpha_{p,t} $ then determine how much each time step contributes to the pooled representation. The resulting context vector $\mathbf{C}_p$ is obtained by taking the attention‑weighted sum of the value projections as :
\begin{equation}
    \mathbf{C}_p = \sum_{t=1}^{T} \alpha_{p,t} \cdot (\mathbf{G}_{p,t} \mathbf{W}_V^p).
    \label{eq:context_vec}
\end{equation}
Finally, the  mass and  friction predictions ( $\widehat{m}$ and $\widehat{\mu}$) are produced by stream-specific MLPs applied to the corresponding context vectors, with output activations chosen to enforce valid physical ranges:
\begin{equation}
    \widehat{m} = \text{Softplus}(\text{MLP}_m(\mathbf{C}_m)) \text{, }  \widehat{\mu} =\text{Sigmoid}(\text{MLP}_\mu(\mathbf{C}_\mu)).
    \label{eq:prop_est}
\end{equation}

\subsection{Physics-Guided Loss Formulation}
Standard supervised loss, often referred to as data loss, allows a neural network to map input-output patterns by minimizing the discrepancy between predictions and ground-truth labels. While effective within the training distribution, such losses are sensitive to real‑world noise and often fail to generalize to unseen conditions because the model learns correlations rather than underlying physical principles~\cite{cai_physics-informed_2021}.
Physics‑guided loss terms address this limitation by constraining the network with fundamental physical equations, encouraging solutions that remain consistent with the true dynamics even outside the training set. 
Consequently, models trained with physics-based constraints are typically more robust to measurement noise and generalize more effectively to novel scenarios~\cite{daw_physics-guided_2022}.

Our physics-guided losses are grounded in Newton's second law and Coulomb's law of friction. Newton's second law states that the acceleration of an object is directly proportional to the net force $F_{\text{net}}$ acting upon it and inversely proportional to its mass, i.e., $F_{\text{net}} = ma$. 
\Done
Coulomb friction, a commonly used model in robotic manipulation, describes the relationship between the tangential friction force $F_{\text{fric}}$ and the normal force $F_{\text{norm}}$.
% via the inequality $F_{\text{fric}} \leq \mu F_{\text{norm}}$, where $\mu$ denotes the friction coefficient. 
In cases where the contact is sliding or experiencing incipient slip, this relationship simplifies to the equality $F_{\text{fric}} = \mu F_{\text{norm}}$. In this regime, the friction force acts opposite to the direction of motion and remains independent of the sliding velocity. We specifically consider this sliding case for our property estimation. \Done To formalize the physics-guided loss, we define the following parameters obtained from simulation. 
\begin{itemize}
    % \item \textbf{Ground truth properties:} $m_{\text{gt}}$ (mass) and $\mu_{\text{gt}}$ (kinetic friction coefficient).
    \item \textbf{Object dynamics:} $a_{\text{obj}}$ (object CoM acceleration), $v_{\text{obj}}$ (object CoM velocity), $F_{\text{net}}$ (net force from robot push force $F_{\text{push}}$ and table friction force $F_{\text{fric}}$), and $F_{\text{norm}}$ (normal force).
    \item \textbf{Dynamic Masks:} $M_{\text{net}, t}, M_{\text{fric}, t} \in \{0, 1\}$ (activity filter masks) with $N_{\text{net}}$ and $N_{\text{fric}}$ representing the total active frames.
    % \item \textbf{Constants \& Hyperparameters:} $g = 9.81$ m/s$^2$ (gravity), $\epsilon = 1 \times 10^{-6}$ for numerical stability, and $\lambda$ coefficients representing the scale weights for mass-related ($\lambda_m$) and friction-related ($\lambda_{\mu}$) loss components.
    \item \textbf{Constants \& Hyperparameters:} $g = 9.81$ m/s$^2$ (gravity) and $\lambda$ coefficients representing the scale weights for mass-related ($\lambda_m$) and friction-related ($\lambda_{\mu}$) loss components.
    % \item \textbf{Loss Criterion:} $\mathcal{L}(\text{target}, \text{theory})$ represents the element-wise loss function.
\end{itemize}
\Update
Our physics-guided loss consists of three variants: the \textit{Independent Force Consistency Loss} $\mathcal{L}_{\text{force}}$, the \textit{Joint Force Consistency Loss $\mathcal{L}_{\text{force, joint}}$}, and the \textit{Force-Acceleration Consistency Loss $\mathcal{L}_{\text{force, acc}}$}.
\Done

% \subsubsection{Semi-Supervised Force-Based Loss $\mathcal{L}_{\text{force}}$}
\subsubsection{\Update Independent Force Consistency Loss $\mathcal{L}_{\text{force}}$ \Done}
This formulation evaluates the force equilibrium for the net force $F_{\text{net}}$ and friction force $F_{\text{fric}}$. For $t \in [T_1, T_2]$, we enforce Newton's Second Law during the transient impact window:
\begin{equation}
    \mathcal{L}_{\text{net}} = \frac{1}{N_{\text{net}}} \sum_{t=T_1}^{T_2} M_{\text{net}, t} \cdot \left| F_{\text{net}, t} - \widehat{m} \cdot a_{\text{obj}}\right|. 
\end{equation}
For $t \in [T_2, T_3]$, we enforce the Coulomb friction equation during the steady-state sliding window, 
\begin{equation}
    \mathcal{L}_{\text{fric}} = \frac{1}{N_{\text{fric}}} \sum_{t=T_2}^{T_3} M_{\text{fric}, t} \cdot \left|F_{\text{fric}, t}- \widehat{\mu} \cdot F_{\text{norm}} \right|.
\end{equation}
The total force-based loss is defined as 
\begin{equation}
    \mathcal{L}_{\text{force}} = \lambda_{m, \text{force}} \mathcal{L}_{\text{net}} + \lambda_{\mu, \text{force}} \mathcal{L}_{\text{fric}}.
    \label{eq:semi_force}
\end{equation}

\subsubsection{\Update Joint Force Consistency Loss $\mathcal{L}_{\text{force, joint}}$ \Done}

% This formulation extends the semi-supervised force equilibrium by introducing a fully unsupervised physical constraint that enforces mathematical consistency between the two separate estimations, $\widehat{m}$ and $\widehat{\mu}$. 

\Update
This formulation extends the independent force equilibrium by introducing a joint physical constraint that enforces mathematical consistency between the two separate estimations, $\widehat{m}$ and $\widehat{\mu}$. 
\Done
This constraint is scaled by a dynamic annealing coefficient $\gamma(E)$ to facilitate a smooth introduction during training. 

\begin{table*}[tb]
\centering
\setlength{\tabcolsep}{3.5pt}
\caption{
Simulation NRMSE(std) ($\downarrow$) for mass ($m$) and friction ($\mu$) across seen and unseen domains. 
\textbf{\xmark} requires force/torque input; \textbf{\cmark} uses only kinematic velocity.}

\begin{tabular}{l|c|c|cccc|c|cccc}
\toprule
\multirow{2}{*}{\textbf{Model}} &
\multirow{2}{*}{\textbf{Input}} &
\multicolumn{5}{|c|}{\textbf{Mass} $m$}  &
\multicolumn{5}{|c}{\textbf{Friction} $\mu$}  \\
\cline{3-12}
 & & $\mathcal{D}_{test}$& $\mathcal{D}_{OOD}^m$ & $\mathcal{D}_{OOD}^{\mu}$ & $\mathcal{D}_{OOD}^{m,\mu}$ & \Update $\mathcal{D}_{OOD}$ \Done & $\mathcal{D}_{test}$& $\mathcal{D}_{OOD}^m$ & $\mathcal{D}_{OOD}^{\mu}$ & $\mathcal{D}_{OOD}^{m,\mu}$ & \Update $\mathcal{D}_{OOD}$ \Done\\
 \midrule

\Update MORRF~\cite{mavrakis_estimating_2020} \Done & \xmark &
\textbf{.004}(.003) & .217(.104) & .045(.026) & .220(.104) & .161(.078) & \textbf{.013}(.011) & \textbf{.094}(.058) & \textbf{.149}(.076) & .318(.104) & .187(.079)\\
\midrule

{PhyPush ($\mathcal{L}_{\text{data}}$)} & {\textbf{\cmark}} &  .044(.030) & .223(.085) & .038(.023) & .212(.091) & .158(.066) &  .093(.070) & .128(.080) & .242(.113) & .258(.065) & .209(.086)\\

{PhyPush ($\mathcal{L}_{\text{force}}$)} & {\textbf{\cmark}}  & .024(.016) & \textbf{.207}(.095) & .034(.021) & \textbf{.198}(.098) & \textbf{.146}(.072)&.054(.042) & .164(.098) & .168(.094) & \textbf{.128}(.067) & .153(.086)\\
% \midrule

{PhyPush ($\mathcal{L}_{\text{force,un}}$)} & {\textbf{\cmark}} & .023(.016) & .209(.097) & \textbf{.034}(.021) & .203(.097) & .149(.072) & .051(.039) & .170(.107) & .158(.090) & .153(.071) & .160(.089)\\

{PhyPush ($\mathcal{L}_{\text{force,acc un}}$)} & {\textbf{\cmark}} & .031(.020) & .219(.097) & .038(.021) & .206(.093) & .154(.070) &.074(.057) & .143(.091) & .170(.092) & .136(.064) & \textbf{.150}(.082)\\

\bottomrule
\end{tabular}
\label{tab:sim_nrmse}
\end{table*}

Let $E$ denote the current training epoch, $E_{\text{start}}$ represent the epoch at which annealing begins, and $D_{\text{ramp}}$ be the duration of the linear fade-in period. The annealing coefficient is defined as s
\begin{equation}
    \gamma(E) = 
    \begin{cases} 
    0, & \text{if } E < E_{\text{start}}, \\
    \min\left(1.0, \frac{E - E_{\text{start}}}{D_{\text{ramp}}}\right), & \text{if } E \geq E_{\text{start}}.
    \end{cases}
\end{equation}

% For the steady-state phase $t \in [T_2, T_3]$, we calculate an unsupervised friction loss using the estimated parameters $\widehat{m}$ and $\widehat{\mu}$, thereby deeply coupling the two predictions 
% \begin{equation}
%     \mathcal{L}_{\text{fric, un}} = \frac{1}{N_{\text{fric}}} \sum_{t=T_2}^{T_3} M_{\text{fric}, t} \cdot \left| F_{\text{fric}, t} - \widehat{\mu} \cdot \widehat{m} \cdot g \right|. 
% \end{equation}
% The total unsupervised force-based loss is defined as the combination of the semi-supervised components and the annealed unsupervised friction term 
% \begin{equation}
%     \mathcal{L}_{\text{force, un}} = \mathcal{L}_{\text{force}} + \gamma(E) \cdot \lambda_{\mu, \text{force}} \mathcal{L}_{\text{fric, un}}.
% \end{equation}

\Update
For the steady-state phase $t \in [T_2, T_3]$, we calculate a joint friction loss using the estimated parameters $\widehat{m}$ and $\widehat{\mu}$, thereby deeply coupling the two predictions 
\begin{equation}
    \mathcal{L}_{\text{fric, joint}} = \frac{1}{N_{\text{fric}}} \sum_{t=T_2}^{T_3} M_{\text{fric}, t} \cdot \left| F_{\text{fric}, t} - \widehat{\mu} \cdot \widehat{m} \cdot g \right|. 
\end{equation}
The total joint force consistency loss is defined as the combination of the independent components and the annealed joint friction term
\begin{equation}
    \mathcal{L}_{\text{force, joint}} = \mathcal{L}_{\text{force}} + \gamma(E) \cdot \lambda_{\mu, \text{force}} \mathcal{L}_{\text{fric, joint}}.
\end{equation}
\Done

% \subsubsection{Unsupervised Acceleration-Based Loss $\mathcal{L}_{\text{acc, un}}$}
% Similarly, this variant acts as an extension to the semi-supervised kinematic constraint $\mathcal{L}_{\text{acc}}$. The unsupervised kinematic consistency is evaluated by computing the theoretical acceleration using only the estimated parameters $\widehat{m}$ and $\widehat{\mu}$ for coupling the property estimations:
% \begin{equation}
%     \mathcal{L}_{\text{cons, un}} = \frac{1}{N_{\text{fric}}} \sum_{t=T_2}^{T_3} M_{\text{fric}, t} \cdot \left| a_{\text{obj}} - \frac{F_{x, t} - (\widehat{\mu} \cdot \widehat{m} \cdot g)}{\widehat{m}} \right|
% \end{equation}
% Likewise, mirroring the structure of the unsupervised force-based loss, the total unsupervised acceleration-based loss is defined as the combination of the semi-supervised components and the annealed unsupervised kinematic consistency term:
% \begin{equation}
%     \mathcal{L}_{\text{acc, un}} = \mathcal{L}_{\text{acc}} + \gamma(E) \cdot \lambda_{\mu, \text{acc}} \mathcal{L}_{\text{cons, un}}
% \end{equation}

% \Done

% \subsubsection{Unsupervised Force/Acceleration-Based Loss $\mathcal{L}_{\text{force, acc, un}}$}
\subsubsection{\Update Force-Acceleration Consistency Loss $\mathcal{L}_{\text{force, acc}}$ \Done}
% Unlike the other two unsupervised losses, $\mathcal{L}_{\text{force, un}}$ and $\mathcal{L}_{\text{acc, un}}$, which are optimized through a single modality (either force or acceleration), this hybrid formulation bridges both domains. 
This is a hybrid formulation that bridges force and acceleration modalities.
% It starts from the semi-supervised force-based loss $\mathcal{L}_{\text{force}}$, where the estimations $\widehat{m}$ and $\widehat{\mu}$ are evaluated in isolation. 
% To enforce deep mathematical coupling, it introduces the unsupervised acceleration-based consistency term $\mathcal{L}_{\text{cons, un}}$, which integrates both $\widehat{m}$ and $\widehat{\mu}$ into a single equation, 
% \begin{equation}
%     \mathcal{L}_{\text{cons, un}} = \frac{1}{N_{\text{fric}}} \sum_{t=T_2}^{T_3} M_{\text{fric}, t} \cdot \left| a_{\text{obj}} - \frac{F_{\text{push}, t} - (\widehat{\mu} \cdot \widehat{m} \cdot g)}{\widehat{m}} \right|.
% \end{equation}
% Similar to the unsupervised force-based loss $\mathcal{L}_{\text{force, un}}$, this term is smoothly incorporated using the curriculum annealing coefficient $\gamma(E)$, 
% \begin{equation}
%     \mathcal{L}_{\text{force, acc, un}} = \mathcal{L}_{\text{force}} + \gamma(E) \cdot \lambda_{\mu, \text{acc}} \mathcal{L}_{\text{cons, un}}.
% \end{equation}

\Update
It starts from the independent force consistency loss $\mathcal{L}_{\text{force}}$, where the estimations $\widehat{m}$ and $\widehat{\mu}$ are evaluated in isolation. 
To enforce deep mathematical coupling, it introduces the joint acceleration consistency term $\mathcal{L}_{\text{acc}}$, which integrates both $\widehat{m}$ and $\widehat{\mu}$ into a single equation, 
\begin{equation}
    \mathcal{L}_{\text{acc}} = \frac{1}{N_{\text{fric}}} \sum_{t=T_2}^{T_3} M_{\text{fric}, t} \cdot \left| a_{\text{obj}} - \frac{F_{\text{push}, t} - (\widehat{\mu} \cdot \widehat{m} \cdot g)}{\widehat{m}} \right|.
\end{equation}
Similar to the joint force consistency loss $\mathcal{L}_{\text{force}}$, this term is smoothly incorporated using the curriculum annealing coefficient $\gamma(E)$, 
\begin{equation}
    \mathcal{L}_{\text{force, acc}} = \mathcal{L}_{\text{force}} + \gamma(E) \cdot \lambda_{\mu, \text{acc}} \mathcal{L}_{\text{acc}}.
\end{equation}
\Done

% The unsupervised kinematic consistency is evaluated by computing the theoretical acceleration using only the estimated parameters $\widehat{m}$ and $\widehat{\mu}$ for coupling the property estimations:
% \begin{equation}
%     \mathcal{L}_{\text{cons, un}} = \frac{1}{N_{\text{fric}}} \sum_{t=T_2}^{T_3} M_{\text{fric}, t} \cdot \left| a_{\text{obj}} - \frac{F_{x, t} - (\widehat{\mu} \cdot \widehat{m} \cdot g)}{\widehat{m}} \right|
% \end{equation}

% \begin{equation}
%     \mathcal{L}_{\text{cons, un}} = \frac{1}{N_{\text{fric}}} \sum_{t=T_2}^{T_3} M_{\text{fric}, t} \cdot \left| a_{\text{obj}} - \frac{F_{x, t} - (\widehat{\mu} \cdot \widehat{m} \cdot g)}{\widehat{m}} \right|
% \end{equation}

\section{Experimental Result}

\subsection{Dataset}
We created a simulated tabletop environment in Isaac Lab \cite{mittal_isaac_2025} (see Fig.\ref{fig:pipeline}), in which a Franka 
% Panda 
robot pushed a $[15\times15\times15]~\text{cm}^{3}$ cube. In this setup, we  varied the cube’s mass $m$ and friction coefficient $\mu$, and collected 31,759 pushing samples. The cube parameters were sampled uniformly from $m$ $\in$ $[0.2, 3.0]$~kg and $\mu$ $\in$ $[0.15, 0.7]$. Data collection was performed in 8,192 parallel environments, with each robot executing a single push from a velocity command $v^{\text{ee, cmd}}$. 
%\Update
% From the pre-push position $p^{\text{pushset}}$ until the object came to a complete rest (, 
For each rollout, we recorded, in five random seeds, the sequence of measured end-effector velocities, $v^{\text{ee, meas}}$, object acceleration $a_{\text{obj}}$, net force $F_{\text{net}}$, robot push force $F_{\text{push}}$, friction force $F_{\text{fric}}$, and normal force $F_{\text{norm}}$, for $t \in [T1, T3]$. The  full dataset was generated in approximately 5 minutes on an NVIDIA GeForce RTX 5070 Ti GPU.

Considering  the parameter ranges $m \in [0.2, 2.0]$~kg and $\mu \in [0.15, 0.5]$, we defined a training set $\mathcal{D}_{train}$ of 12,550 samples and an in-distribution test set $\mathcal{D}_{test}$ (also \textbf {Seen} domain) of 3,138 samples. 
To further assess the capability and generalization performance of PhyPush, we also created an \textbf{ Unseen} domain with out-of-distribution test sets $\mathcal{D}_{OOD}$ (16,071 samples)  defined as follows:
% \begin{itemize}
%     \item $\mathcal{D}_{OOD}^m$ (8,436 samples), with  $m \in [0.1, 0.3[ \, \cup \, ]1.0, 2.0]$ and $\mu \in [0.3, 0.5]$
%     \item $\mathcal{D}_{OOD}^{\mu}$ (4,641 samples), with  $\mu \in [0.2, 0.3[ \,\cup\, ]0.5, 0.6]$ and $m \in [0.3, 1.0]$
%      \item $\mathcal{D}_{OOD}^{m, \mu}$ (7,843 samples), with $m \in [0.1, 0.3[ \,\cup\, ]1.0, 2.0]$ and $\mu \in [0.2, 0.3[ \, \cup \,]0.5, 0.6]$
% \end{itemize}
\begin{itemize}
    \item $\mathcal{D}_{OOD}^m$ 8,589 samples with  $m {\in} [2.0, 3.0]$, $\mu {\in} [0.15, 0.5]$,
    \item $\mathcal{D}_{OOD}^{\mu}$ 4,865 samples,with  $\mu {\in} [0.5, 0.7]$, $m {\in} [0.2, 2.0]$,
     \item $\mathcal{D}_{OOD}^{m, \mu}$ 2,617 samples with $m {\in} [2.0, 3.0]$, $\mu {\in} [0.5, 0.7]$.
\end{itemize}

\subsection{Training setup and evaluation metrics}
PhyPush used a 4-layer Transformer encoder, where each layer is configured with 4 multi-head attention streams.
For the property-specific queries, we employed attention sharpness hyperparameters of $s_{m} = 5.0$ and $s_{\mu} = 5.0$.
The model was trained for 1,000 epochs using the AdamW optimizer with a batch size of 64. 
We utilized a one-cycle learning rate scheduler with an initial learning rate of $3 \times 10^{-4}$.
% For the unsupervised loss variants $\mathcal{L}_{\text{force, un}}$ and $\mathcal{L}_{\text{force, acc, un}}$, 
\Update
For the joint consistency loss variants $\mathcal{L}_{\text{force, joint}}$ and $\mathcal{L}_{\text{force, acc}}$, 
\Done
we implemented an annealing schedule defined by a start epoch $E_{\text{start}} = 300$ and a ramp-up duration $D_{\text{ramp}} = 600$. 
The loss weighting coefficients were empirically determined and set to $\lambda_{m, \text{force}} = 7.5$, $\lambda_{\mu, \text{force}} = 1.0$, and $\lambda_{\mu, \text{acc}} = 1.0$. 
To ensure the physical validity of the training signals, the dynamic masks $M_{\text{net}, t}$ and $M_{\text{fric}, t}$ were used to filter out noise, specifically excluding frames where $a_{\text{obj}} < 0.3$~m/s$^2$ and $v_{\text{obj}} < 0.01$~m/s, respectively. 

We recorded the velocity input using a window of size T=60. PhyPush was then evaluated with the Normalized Root-Mean-Square Error (NRMSE) and the raw bias metric for both mass and friction estimations, enabling performance comparisons across different scales. \cite{mavrakis_estimating_2020}.
%This metric is particularly sensitive to large outliers, ensuring that rare but significant estimation failures are accounted for.
% \begin{equation}
%     \text{NRMSE} = \left( \frac{\sqrt{\frac{1}{n} \sum_{i=1}^{n} (y_i - \hat{y}_i)^2}}{y_{\text{max}} - y_{\text{min}}} \right)
%     \label{nrmse}
% \end{equation}

\subsection{Simulation Results}
We validated our framework on the 7 DoF Franka Emika  robot performing an end-effector guided push in both simulation and real‑world settings. 
% PhyPush was compared against a baseline, state-of-the-art method \cite{mavrakis_estimating_2020} that relies on a force sensor.
\Update
PhyPush was compared against a state-of-the-art baseline method \cite{mavrakis_estimating_2020} employing a Multi-Output Random Forest Regressor (MORRF), which relies on force sensor data and object velocity in addition to end-effector velocity.
\Done
We re‑implemented this baseline to the best of our ability and adapted it to our translational pushing setup. 
% % 
% Importantly, the baseline relies on privileged information with precise force measurements. 
% % simulation we provided it with precise force measurement}
% % 
% 
\Update
Importantly, the baseline relies on privileged information with precise force measurements and object velocity.
\Done
In addition to the baseline, 
% we evaluated PhyPush with its variants' physics‑guided losses ($\mathcal{L}_{\text{force}}$, $\mathcal{L}_{\text{force, un}}$, $\mathcal{L}_{\text{force, acc, un}}$), 
\Update
we evaluated PhyPush with its variants' physics‑guided losses ($\mathcal{L}_{\text{force}}$, $\mathcal{L}_{\text{force, joint}}$, $\mathcal{L}_{\text{force, acc}}$), 
\Done
as well as a purely data‑driven loss ($\mathcal{L}_{\text{data}}$).

% ($\mathcal{L}_{\text{force}}$, $\mathcal{L}_{\text{acc}}$, $\mathcal{L}_{\text{force, un}}$, $\mathcal{L}_{\text{acc, un}}$, $\mathcal{L}_{\text{force, acc, un}}$), as well as a purely data‑driven loss ($\mathcal{L}_{\text{data}}$).

\Done
Table \ref{tab:sim_nrmse} summarizes the NRMSE performance of all models across the different evaluation domains.
While the baseline \cite{mavrakis_estimating_2020} achieved the lowest errors on the in‑domain set $\mathcal{D}_{test}$, 
% the PhyPush variants consistently outperformed it on all out‑of‑distribution domains sets $\mathcal{D}_{OOD}$ 
\Update the PhyPush with physics-guided loss variants consistently outperformed it on overall out‑of‑distribution domains sets $\mathcal{D}_{OOD}$ \Done
despite requiring no force input. 
\Update
This is particularly notable in the most challenging setting $\mathcal{D}_{OOD}^{m, \mu}$, where PhyPush ($\mathcal{L}_{\text{force}}$) achieves lower estimation errors for both mass $m$ ($\downarrow$10\%) and friction $\mu$ ($\downarrow$60\%) using only the end‑effector velocity.
\Done
These results highlight the strong generalization and real-world applicability of the PhyPush architecture. Crucially, the comparison between loss functions shows that physics‑guided losses consistently outperform the data‑only loss, which is a central contribution of this work. 
For instance, the variant, \Update $\mathcal{L}_{\text{force, acc}}$ \Done, reduces the friction‑estimation error by more than 28\% relative to the data‑only loss $\mathcal{L}_{\text{data}}$, highlighting the clear benefit of embedding physical constraints directly into the learning objective.

% In this domain, the baseline and $\mathcal{L}_{\text{data}}$ predicted $\mu$ with 30\% and 25\% errors, respectively.
% In contrast, PhyPush trained with $\mathcal{L}_{\text{force}}$ achieved only 10\% error, emphasizing the model’s physics fidelity and generalizability.

\Done

\subsection{PhyPush Insights}

\Done

% Ablation study\\
% Physics law consistency\\
% Attention?

To evaluate the benefits of co-estimation, we compared the performance of our joint estimator (PhyPush trained with $\mathcal{L}_{\text{force}}$) against baselines trained to estimate only mass ($m$) or only friction ($\mu$), as summarized in Table~\ref{tab:co_est}. 
Our co-estimator consistently outperforms both single-property estimators on the in-domain test set $\mathcal{D}_{\text{test}}$ and the out-of-distribution set $\mathcal{D}_{\text{OOD}}$. 
This highlights the inherent advantage of joint optimization, suggesting that simultaneously estimating $m$ and $\mu$ yields a more robust and physically meaningful representation than predicting them in isolation.

\begin{table}[tbp]
    \centering \setlength{\tabcolsep}{2pt}
    \caption{NRMSE(std) ($\downarrow$)  for different training conditions (Cond.) of PhyPush for Seen and Unseen domains.}
    \label{tab:co_est}
    \begin{tabular}{l| ccc| ccc }
        \toprule
        \multirow{2}{*}{\textbf{Cond.}} & \multicolumn{3}{|c|}{\textbf{Mass} $m$} & \multicolumn{3}{|c}{\textbf{Friction} $\mu$} \\
       \cline{2-7}
         & $\mathcal{D}_{test}$ & $\mathcal{D}_{OOD}$ & Overall & $\mathcal{D}_{test}$ & $\mathcal{D}_{OOD}$ & Overall\\
         \midrule
        {$m$ only} & .031(.020) & .152(.071) & .092(.046) & - & - & - \\
        {$\mu$ only} & - & - & - & .106(.065) & .130(.064) & .118(.065)\\
        {$\mathcal{L}_{\text{force}}$} & \textbf{.024}(.016) & \textbf{.146}(.072) & \textbf{.085}(.044) & \textbf{.054}(.042) & \textbf{.153}(.086) & \textbf{.104}(.064)\\
        \bottomrule
    \end{tabular}
\end{table}

\begin{table}[tbp]
    \centering \setlength{\tabcolsep}{2pt}
    % \caption{Physics law consistency, $R^2$ Scores ($\uparrow$), across domains for PhyPush trained with data loss ($\mathcal{L}_{\text{data}}$) and physics-guided loss ($\mathcal{L}_{\text{force,acc un}}$)}
    \caption{\Update Physics law alignment, $R^2$ Scores ($\uparrow$), across domains for PhyPush trained with data loss ($\mathcal{L}_{\text{data}}$) and physics-guided loss ($\mathcal{L}_{\text{force, acc}}$) \Done}
    \label{tab:phy_law_cons}
    \begin{tabular}{l c c c c c c}
        \toprule
        \textbf{Model} & \textbf{Law} & $\mathcal{D}_{test}$ & $\mathcal{D}_{OOD}^m$ & $\mathcal{D}_{OOD}^{\mu}$ & $\mathcal{D}_{OOD}^{m, \mu}$ & Overall \\
        \midrule
        \multirow{2}{*}{ $\mathcal{L}_{\text{data}}$} & $F_{\text{net}}=\widehat{m}a_{\text{obj}}$ & 0.95 & -3.12 & 0.96 & -2.89 & -1.03 \\
        \cmidrule(lr){2-7}
         & $F_{\text{fric}}=\widehat{\mu}\widehat{m}g$ & 0.91 & 0.15 & 0.75 & -7.17 & -1.34 \\
        \midrule
        \multirow{2}{*}{$\mathcal{L}_{\text{force,acc un}}$} & $F_{\text{net}}=\widehat{m}a_{\text{obj}}$ & \textbf{0.97} & \textbf{-2.97} & \textbf{0.96} & \textbf{-2.68} & \textbf{-0.93} \\
        \cmidrule(lr){2-7}
         & $F_{\text{fric}}=\widehat{\mu}\widehat{m}g$ & \textbf{0.98} & \textbf{0.58} & \textbf{0.93} & \textbf{-4.12} & \textbf{-0.41} \\
        \bottomrule
    \end{tabular}
\end{table}

% Moreover, we verified the consistency of the estimations with fundamental physical laws by substituting the predicted mass $\widehat{m}$ and friction coefficient $\widehat{\mu}$ into Newton's Second Law and Coulomb's Law of Friction, as detailed in Table~\ref{tab:phy_law_cons}. 
% We used the coefficient of determination ($R^2$) to evaluate the physical consistency of our model's outputs. Specifically, we measured how closely the predicted inertial term ($\widehat{m}a_{\text{obj}}$) aligns with the observed net force ($F_{\text{net}}$) according to Newton's Second Law. 
\Update
Moreover, we verified the alignment of the estimations with fundamental physical laws by substituting the predicted mass $\widehat{m}$ and friction coefficient $\widehat{\mu}$ into Newton's Second Law and Coulomb's Law of Friction, as detailed in Table~\ref{tab:phy_law_cons}. 
Specifically, we measured how closely the predicted inertial term ($\widehat{m}a_{\text{obj}}$) aligns with the observed net force ($F_{\text{net}}$) and how the predicted friction($\widehat{m}\widehat{\mu}g$) aligns with the observed friction force ($F_{\text{fric}}$). We used the coefficient of determination ($R^2$) to evaluate the physical alignment of our model’s outputs. A value $R^2$  closer to 1 suggests perfect consistency.
\Done PhyPush trained with the physics-guided loss 
% $\mathcal{L}_{\text{force, acc un}}$
\Update $\mathcal{L}_{\text{force, acc}}$ \Done
consistently outperformed the purely data-driven baseline $\mathcal{L}_{\text{data}}$ across all tested domains. Notably, when evaluating on the unseen mass domain ($\mathcal{D}_{\text{OOD}}^m$), the baseline's friction estimation nearly collapses (0.153), whereas the physics-guided model recovers performance significantly (0.579). Similarly, when evaluated on the unseen friction domain ($\mathcal{D}_{\text{OOD}}^{\mu}$), the baseline shows degraded performance (0.757), while the proposed model maintains excellent accuracy (0.933), performing almost on par with its in-domain results. \Update Nonetheless, law-consistency remains imperfect under the most severe compound shift $\mathcal{D}_{OOD}^{m, \mu}$, where $R^2$ stays negative for both losses.  While physics-guided losses consistently improve law-consistency, residual error remains non-negligible under large or compound distribution shifts, indicating that physics-guided supervision mitigates but do not  fully close the consistency gap in the most challenging regimes. \Done
% This clearly demonstrates the benefit of our curriculum-style annealing and the deep coupling of $\widehat{m}$ and $\widehat{\mu}$ in the unsupervised loss 
% $\mathcal{L}_{\text{force, acc un}}$.
%\Update $\mathcal{L}_{\text{force, acc}}$. \Done

% Similarly, we assessed the alignment between the product of the estimated parameters ($\widehat{\mu} \widehat{m} g$) and the observed friction force ($F_{\text{fric}}$) to check the model's adherence to the Coulomb friction model.

% \begin{table}[htbp]
%     \centering
%     \caption{Physics Law Consistency ($R^2$ Scores) ($\uparrow$) Across Domains}
%     \label{tab:phy_law_cons}
%     \begin{tabular}{l c c c c c c}
%         \toprule
%         \textbf{Model} & \textbf{Law} & $\mathcal{D}_{test}$ & $\mathcal{D}_{OOD}^m$ & $\mathcal{D}_{OOD}^{\mu}$ & $\mathcal{D}_{OOD}^{m, \mu}$ & Overall \\
%         \midrule
%         \multirow{2}{*}{$\mathcal{L}_{\text{data}}$} & $F_{\text{net}}=\widehat{m}a_{\text{obj}}$ & 0.946 & -3.123 & 0.961 & -2.889 & -1.026 \\
%         \cmidrule(lr){2-7}
%          & $F_{\text{fric}}=\widehat{\mu}\widehat{m}g$ & 0.910 & 0.153 & 0.757 & -7.173 & -1.338 \\
%         \midrule
%         \multirow{2}{*}{$\mathcal{L}_{\text{force,acc un}}$} & $F_{\text{net}}=\widehat{m}a_{\text{obj}}$ & \textbf{0.974} & \textbf{-2.966} & \textbf{0.962} & \textbf{-2.683} & \textbf{-0.928} \\
%         \cmidrule(lr){2-7}
%          & $F_{\text{fric}}=\widehat{\mu}\widehat{m}g$ & \textbf{0.977} & \textbf{0.579} & \textbf{0.933} & \textbf{-4.118} & \textbf{-0.407} \\
%         \bottomrule
%     \end{tabular}
% \end{table}

\Done

\subsection{Real-world Results}

\begin{figure}[t]
    \centering
    \includegraphics[width=0.9\columnwidth]{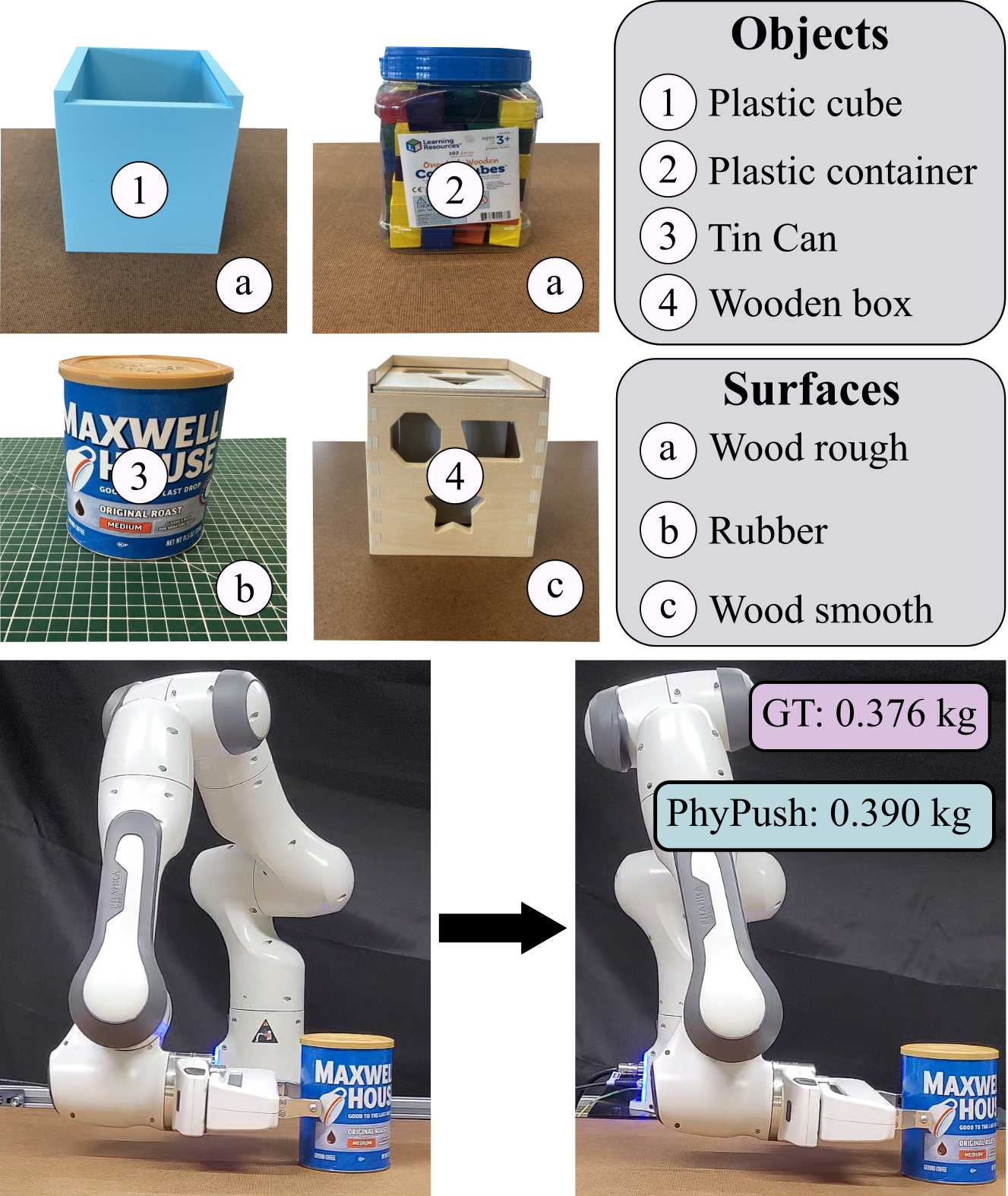}
    \caption{
        \small PhyPush was validated experimentally across objects and surfaces with diverse physical properties.
    }
    \label{fig:real_exp}
\end{figure}
% \Update
We conducted an extensive experimental study across three different surfaces and a range of objects with varying weights, as illustrated in Fig.~\ref{fig:real_exp}.
\Update
Importantly, to evaluate the PhyPush models' ability to bridge the sim-to-real gap, they were deployed purely zero-shot. They were transferred directly from simulation to real robotic hardware without any fine-tuning or retraining.
\Done
These objects included a plastic cube identical to the training setting, and completely novel objects such as a tin can, and a plastic container.  For each object, the mass could be adjusted by adding material inside, enabling a broad range of testing conditions. In total, this resulted in 47 distinct object–surface configurations. 
% \Done
The ground-truth mass, $m_{gt}$, of each real object was measured using a digital scale. 
The ground-truth kinetic friction coefficient, $\mu_{gt}$, was determined using an inclined plane and a camera-based tracking method. 
% Each object was placed on the inclined plane and released from rest, ensuring an initial velocity of zero ($v_0 = 0$). 
% \Update
Each object was placed on an inclined plane at angle $\theta$ and released from rest, with initial velocity $v_0 = 0$. 
% \Done
% The incline angle, $\theta$, was measured with a digital protractor, and the travel distance, $d$, was measured with a measuring tape. 
% 
The elapsed time $t$ required to travel a known distance $d$ along the surface was obtained from video recordings. Assuming that the transient transition from static to kinetic friction is short relative to the overall sliding duration, the subsequent motion can be approximated as uniformly accelerated. Under this assumption, the acceleration 
$a$  is estimated from the standard kinematic relation:  
% 
% As the object slid down the surface, the elapsed time, $t$, required to travel the distance $d$ could be determined from a video recording. 
% Because the transition from static to kinetic friction is brief relative to the total travel time $t$, the object’s motion can be well-approximated as having constant acceleration. Under this assumption, the acceleration $a$ follows directly from the kinematic relation for displacement 
% 
% 
\begin{align}\label{eq:mu_acceleration}
    a = \tfrac{2d}{t^2} \text{ with } d = v_0t + \tfrac{1}{2}at^2.  
\end{align}
To estimate the friction coefficient $\mu_{gt}$, we applied Newton’s Second Law along the direction of motion on the inclined plane, leading to
\begin{equation}\label{eq:mu_geometry}
    mg \sin(\theta) - \mu_{gt} mg \cos(\theta) = ma. 
\end{equation}
\begin{figure*}[t]
    \centering
    \includegraphics[width=1\textwidth]{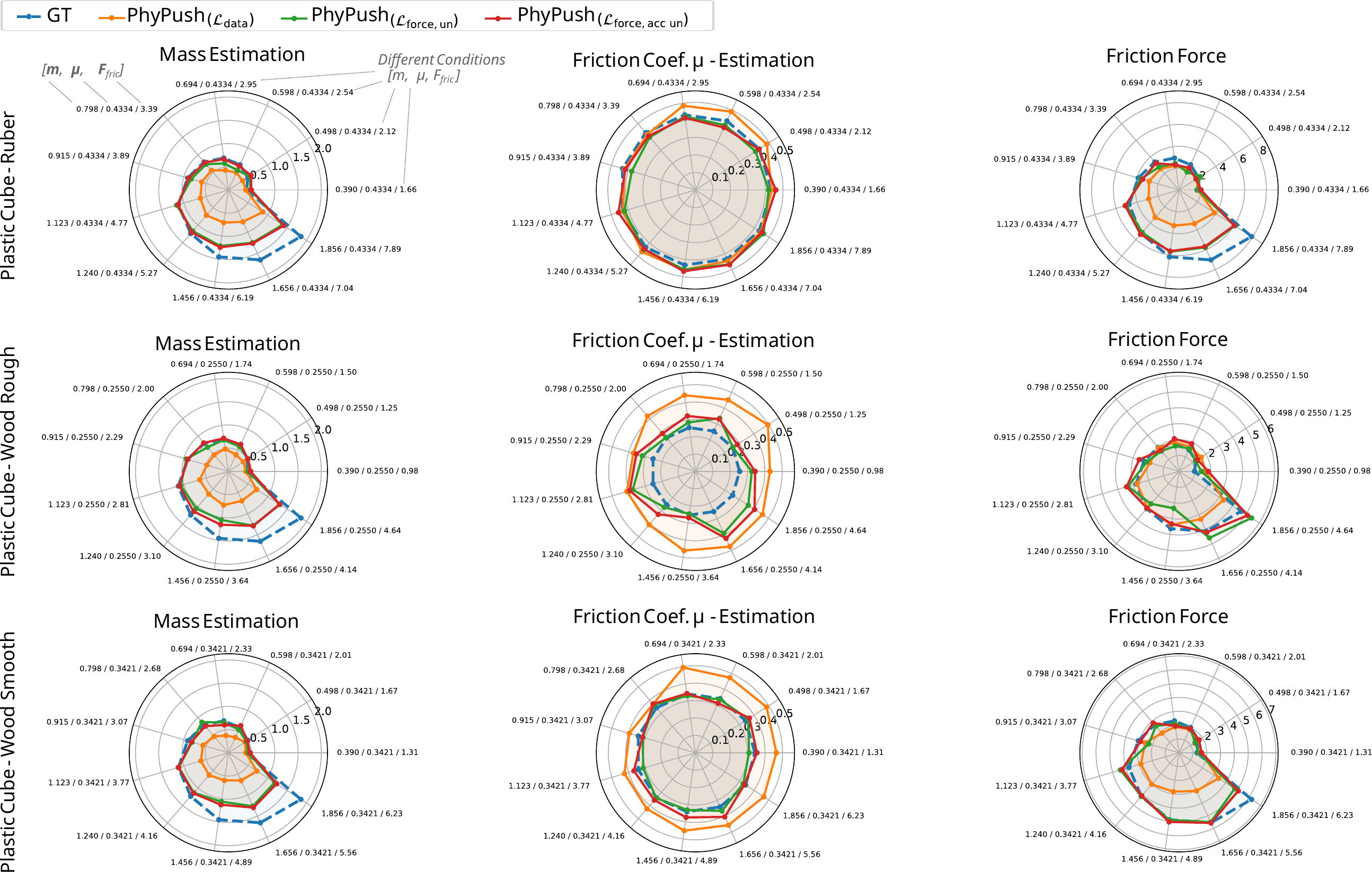}
    \caption{\small 
    Real-world performance of PhyPush on seen objects. Each row corresponds to one of the 33 evaluated object–surface–conditions (plastic cube with rubber, wood rough, and smooth surfaces), and the three columns report mass, friction coefficient, and the derived friction force $F_{\mathrm{fric}}=\hat{\mu}\hat{m}g$. The radar vertices denote the different ground-truth conditions $[m,\mu,F_{\mathrm{fric}}]$, while the curves compare the ground truth (GT in blue) against the three training variants: PhyPush with $\mathcal{L}_{\mathrm{data}}$, %$\mathcal{L}_{\mathrm{force,un}}$, and $\mathcal{L}_{\mathrm{force,acc\,un}}$. 
    \Update $\mathcal{L}_{\mathrm{force,joint}}$, and $\mathcal{L}_{\mathrm{force, acc}}$.  \Done
    Across tests, the physics-guided variants track the ground-truth trends more closely and more consistently than the purely data-driven version, particularly for the physically coupled quantities $\mu$ and $F_{\mathrm{fric}}$.
    }
    \label{fig:results_spider_seen}
\end{figure*}
\begin{figure*}[t]
    \centering
    \includegraphics[width=1\textwidth]{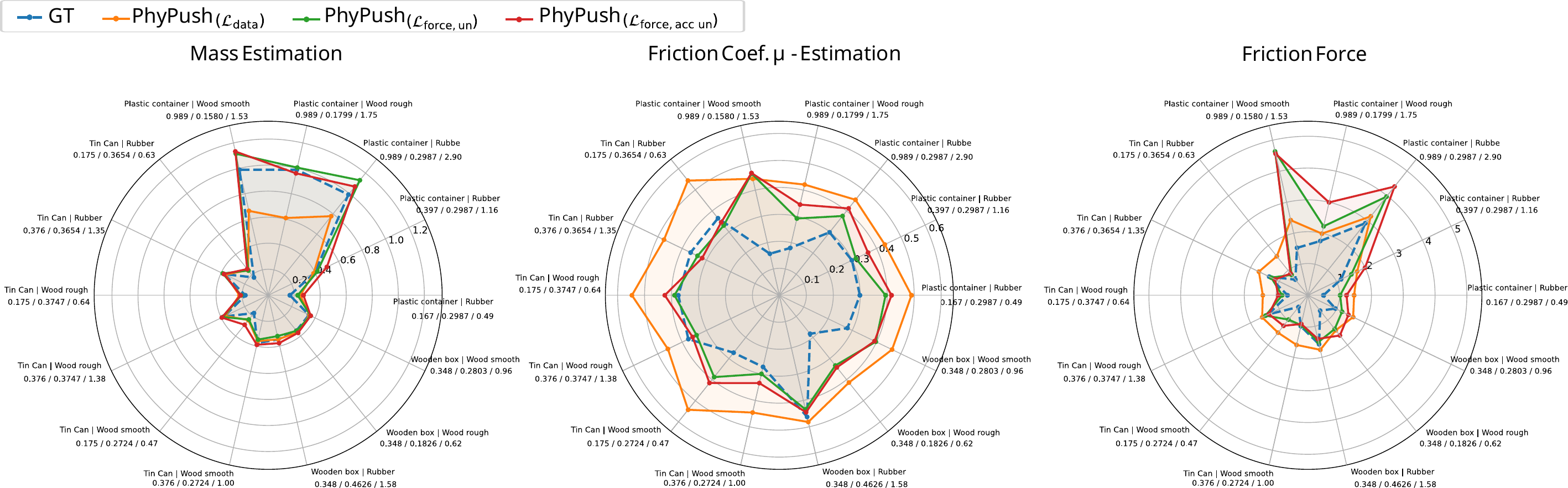}
    \caption{\small 
    Real-world performance of PhyPush on unseen objects across 14 object--surface--condition combinations, including the plastic container, tin can, and wooden box. The three columns report mass, friction coefficient, and the derived friction force $F_{\mathrm{fric}}=\hat{\mu}\hat{m}g$. Each radar vertex corresponds to one unseen test condition, identified by the object, contact surface, and ground-truth tuple $[m,\mu,F_{\mathrm{fric}}]$. The plots compare ground truth (GT, in blue) against the three training variants: 
    PhyPush with $\mathcal{L}_{\mathrm{data}}$, % $\mathcal{L}_{\mathrm{force,un}}$, and $\mathcal{L}_{\mathrm{force,acc\,un}}$. 
    \Update $\mathcal{L}_{\mathrm{force, joint}}$, and $\mathcal{L}_{\mathrm{force, acc}}$. \Done
    Consistent with the simulation and real-world trends reported in the paper, the physics-guided losses improve out-of-distribution behavior and produce more reliable friction-related estimates than the purely data-driven variant.
    }
    \label{fig:results_spider_unseen}
\end{figure*}
By injecting \eqref{eq:mu_acceleration} into  \eqref{eq:mu_geometry}, a closed-form solution of $\mu_{gt}$ is 
\begin{equation} 
    \mu_{gt} = \tan(\theta) - \frac{2d}{g t^2 \cos(\theta)}.
\end{equation}

\Update
For each configuration, the pushing action was repeated ten times. The resulting average estimates for each PhyPush variant, evaluated against ground truth mass, friction and friction force, are summarized in Figs. \ref{fig:results_spider_seen} and \ref{fig:results_spider_unseen}. 

% To further assess the physical fidelity of the predictions, we also compared the average estimated friction force computed as $F_{\text{fric}}=\widehat{\mu}\widehat{m}g$, with the ground truth one.
% The spider plots confirm that PhyPush physics-guided loss variants ($\mathcal{L}_{\text{force, un}}$ and $\mathcal{L}_{\text{force, acc, un}}$) 
\Update
The spider plots confirm that PhyPush physics-guided loss variants ($\mathcal{L}_{\text{force, joint}}$ and $\mathcal{L}_{\text{force, acc}}$)
\Done
consistently outperform the data-driven variant (Fig.~\ref{fig:results_spider_seen})  
across the tested physical parameters and contact conditions.
In the setting of seen objects, the data-driven $\mathcal{L}_{\text{data}}$ exhibits larger deviations and inconsistent scaling. 
A clear pattern of object's mass underestimation and friction coefficient overestimation can be observed.
% Furthermore, in cases where the mass estimation error from $\mathcal{L}_{\text{force, un}}$ and $\mathcal{L}_{\text{force, acc, un}}$ reached around 0.5 kg for heavier objects such as 1.456 kg and 1.656 kg, the total friction force error remained small and sometimes nearly zero.
This shows our model's strong physics law consistency in the real world.

In the more challenging setting of unseen objects, illustrated in Fig.~\ref{fig:results_spider_unseen}, we observed the same performance patterns between the data-driven and physics-guided loss.
Our physics-guided loss variants continue to track the ground-truth trends more closely, despite the fact that these objects were not encountered during training.
% successfully track the mass ground truth, despite the fact that the model has not encountered the specific object during training.
In some systems that use contact sensors, such as a tactile sensor, input signals can become significantly noisier or different when object shapes differ.
In contrast, our physics-guided loss variants' results demonstrate not only the robustness of our model's estimation capabilities but also the practical applicability of PhyPush, which relies on easily accessible robotic end-effector velocity rather than specialized hardware.

\Done

\section{Conclusion}
This work introduced PhyPush, a sensorless framework that estimates an object's mass and friction coefficient from a single push using only end-effector velocity. By exploiting the direct coupling between the robot’s kinematic response and the object’s dynamics, PhyPush removes the need for force/torque sensing while maintaining high accuracy.
At the core of our approach is a Transformer-based architecture trained with a physics‑guided loss derived from Newton’s Second Law and Coulomb friction.  The physics-based constraints enforce mathematical consistency and improve generalization beyond the training distribution.
We extensively demonstrated the effectiveness of PhyPush in simulation by outperforming by more than 10\% a baseline with
full force information \Update in the most challenging setting,  and in real-world experiments by showing that physics-grounded supervision yields more accurate and more robust predictions than purely data-driven losses. \Done

While PhyPush currently assumes predominantly translational pushes and relies on velocity measurements, these constraints open several promising research directions. Incorporating rotational dynamics into the physics-guided loss would enable estimation under richer interaction settings. Additionally, developing an adaptive pushing policy, for example via reinforcement learning, could tailor the interaction to object geometry and generate a broader range of informative physical responses beyond those available under a translational push only.

\bibliographystyle{ieeetr}
\bibliography{sections/PhyPush}
\flushbottom
\end{document}